# AlphaDent: A dataset for automated tooth pathology detection

Evgeniy I. Sosnin[a], Yuriy L. Vasilev[a], Roman A. Solovyev[b], Aleksandr L. Stempkovskiy[b],
Dmitry V. Telpukhov[b], Artem A. Vasilev[b], Aleksandr A. Amerikanov[c], Aleksandr Y. Romanov[c]
[a]Sechenov University, Moscow, Russia
[b]AlphaChip LLC, Moscow, Russia
[c]HSE University, Moscow, Russia

## ABSTRACT

In this article, we present a new unique dataset for dental research – AlphaDent. This dataset is based on the DSLR camera photographs of the teeth of 295 patients and contains over 1200 images. The dataset is labeled for solving the instance segmentation problem and is divided into 9 classes. The article provides a detailed description of the dataset and the labeling format. The article also provides the details of the experiment on neural network training for the Instance Segmentation problem using this dataset. The results obtained show high quality of predictions. The dataset is published under an open license; and the training/inference code and model weights are also available under open licenses.

## 1. INTRODUCTION

The decisive link in high-quality dental care is the accurate diagnostics, diagnosis, and treatment plan. Modern dentistry includes intraoral photo protocol in diagnostics. Currently, the analysis of photographs of the patient's teeth is a labor-intensive and energy-consuming process. Often, the process of detecting certain pathological processes in the oral cavity and dental defects can be complicated due to the doctor's fatigue during the work process, lack of time between the clinical appointments with patients and analysis of photographic data, which significantly complicates and extends the time for drawing up a treatment plan, in turn slowing down the decision-making process and the transition directly to treatment. Artificial intelligence methods can help the doctor in making a diagnosis.

Modern deep learning methods in dentistry require large and carefully labeled datasets to train and validate models. However, the analysis conducted in [1] showed that there are very few publicly available dental datasets - only 16 unique sets (about 10,450 images). Moreover, the vast majority of such data are intended for tooth segmentation (62.5% of sets) or their labeling (56.2%), and 58.8% of images are panoramic radiographs. Less than half of the collections contain information on the ethical review (31.2% with approved consent) and have explicit licensing terms (56.25% of datasets did not indicate a license). Taken together, these data indicate a severe lack of open data in dentistry.

The main types of dental images are radiographs of various types. Panoramic radiography (panoramic images of the jaws) is the most common. There are large open sets of panoramic images, such as the Swiss DENTEX Panoramic (more than 2000 images) [2,3]. Other examples include datasets from the US and Iran – Tufts Panoramic [4] (1000 images), as well as specialized collections with caries markings. Such datasets often publish masks of teeth or cavities for segmentation and classification tasks. For example, for cephalometric radiographs (lateral images of the skull), the Grand-Challenge dataset "CL Detection" (600 images marked with pixel labels for recognizing key points on the jaw) is available [5]. Many X-ray datasets contain annotations of teeth, pathologies, and tissue atrophy. A large list of dental datasets is available at the following link [6].

Among the three-dimensional datasets, the ones based on the cone beam computed tomography (CBCT) stand out.

Large multimodal datasets have been published recently. For example, Wang et al. presented STS-Tooth – a set of 4000 panoramic images with 900 labeled tooth masks and 148400 CBCT scans (8800 with labels [7]. Liu et al. collected a multimodal dataset on PhysioNet, including 574 CBCT images from 389 patients and additional 29,199 periapical (intraoral) radiographs generated from these CBCT images [8]. Huang et al. (Scientific Data, 2024) published a set of 329 CBCT, 8 panoramic and 16,203 periapical images for the same patients [9]. For 3D scans of the oral cavity, there is a dataset 3DTeethSeg22 (ToothFairy) consisting of 1800 3D images (half male/female) with segmentation of teeth [10].

Datasets with intraoral photographs (to which our work relates) deserve special attention. In such datasets, dental surfaces and various objects in the photo (gingiva, plaque, caries, elements of orthodontic structures, etc.) are annotated. For example, in [11,12], the SegmentAnyTooth framework with 5000 photographs of teeth images (953 subjects) with marking of tooth surface contours and indication of FDI numbers is presented. But the dataset itself is not publicly available, and the model weights are available upon request. Researchers in pediatric dentistry collected 886 photographs of the baby teeth with a dye (which detects plaque) and detailed labeling of the plaque and tooth area (annotations were done manually in LabelMe) [13]. Liu et al. use a dataset of 3,365 photos labeled with gingivitis, caries, and tartar (segmentation of the two disease types) [14]. Thus, existing oral imaging datasets typically contain annotations of teeth, gums, plaque, cavities, and even orthodontic components (brackets, wires, etc.), although such publicly available data is still rare.

Despite progress, there are still serious limitations in this area of research: the number of open datasets is limited, many of them are small and focus on a narrow range of tasks. More than half of the datasets do not have a clearly defined license at all [1]. The data are often heterogeneous in formats (DICOM, JPEG, PNG, etc.) and markup standards, which makes them difficult to combine. Some important pathologies or patient groups are practically not represented; for example, only one open dataset for the recognition of oral cancer was published [15]. In addition, less than 2% of dental studies share their images in machine-readable form. Together, this limits the generalizability and robustness of the models: without diverse and standardized data, AI systems may perform poorly for new populations or unfamiliar pathologies. Therefore, consolidating and expanding open dental datasets, standardizing them (including creating "data cards" with metadata), and ensuring FAIR principles are priorities for further development of AI in this field. Only with the availability of high-quality labeled data the medical algorithms will be able to achieve high accuracy and broad applicability in dentistry to improve the diagnosis and treatment of patients.

In this article, we propose the first of its kind fully open dataset based on high-resolution intraoral photographs. The dataset is labeled with masks and designed to solve the Instance Segmentation problem [16]. The markup contains 9 classes. In addition to the dataset, we prepared an open code for training and inference of neural networks; a set of pre-trained weights and Leaderboard with hidden test data labels for fair comparison with solutions of other researchers.

## 2. MATERIAL AND METHODS

To achieve our goal, an anonymous database of real patients' photographs was prepared. In total, 1320 photographs of 295 patients were used in the study (several photographs are available for each patient). The photographs were taken using an intraoral mirror, a Canon 6D mark ll camera, and a Canon 100mm f/2.8l macro is usm lens. The photographs were collected from January 2024 to January 2025. Using the CVAT marking website [17,18], dental hard tissue defects were manually marked by dentists.

## 3. DATA COLLECTION

During the data marking, visually noticeable problem areas in the photograph and hidden carious cavities were assessed based on local signs such as gray enamel and darkening of individual areas of the teeth with localization characteristic of caries. Caries was assessed according to 6 classes (5 standard classes proposed in [19,20] and the 6th class proposed by WHO, with localization on the immune zones of teeth (tubercles of molars, premolars and cutting edges of incisors). Artificial crowns found in the marking were mainly metal-ceramic, metal, ceramic and zirconium. The dataset contains both frontal images and separate images of the chewing surface for the upper and lower jaw. Each tooth was assessed in detail for the presence of wear facets, both pathological and physiological. A total of 9 classes were marked as shown in Table 1.

Table 1. Pathology classes in the dataset

| Class ID | Name | Description |
|---|---|---|
| 0 | Abrasion | Teeth with mechanical wear of hard tissues. |

| 1 | Filling | Fillings of various types. |
|---|---|---|
| 2 | Crown | Installed crown. |
| 3 | Caries 1 class | Caries in fissures and blind pits of teeth (occlusal surfaces of molars and premolars, buccal surfaces of molars, lingual surfaces of upper incisors). |
| 4 | Caries 2 class | Caries of the contact surfaces of molars and premolars. |
| 5 | Caries 3 class | Caries of the contact surfaces of incisors and canines without damage to the cutting edge. |
| 6 | Caries 4 class | Caries of the contact surfaces of incisors and canines with damage to the cutting edge. |
| 7 | Caries 5 class | Cervical caries of the vestibular and lingual surfaces. |
| 8 | Caries 6 class | Caries of the cutting edges of the front teeth and the cusps of the chewing teeth. |

The dataset is primarily intended to solve the Instance Segmentation problem [16], therefore, in addition to the bounding boxes, carefully marked masks are available for each pathology. Examples of the marked images are shown in Fig. 1.

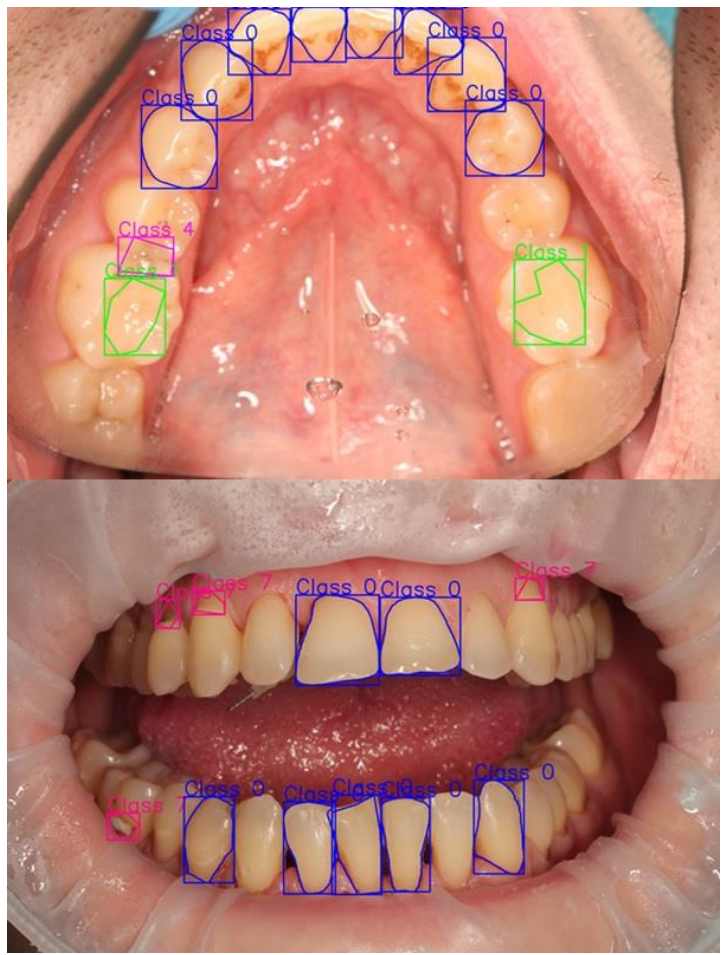

Figure 1. Examples of photographs of the oral cavity with markings

Figure 2 shows the examples of markup for each class.

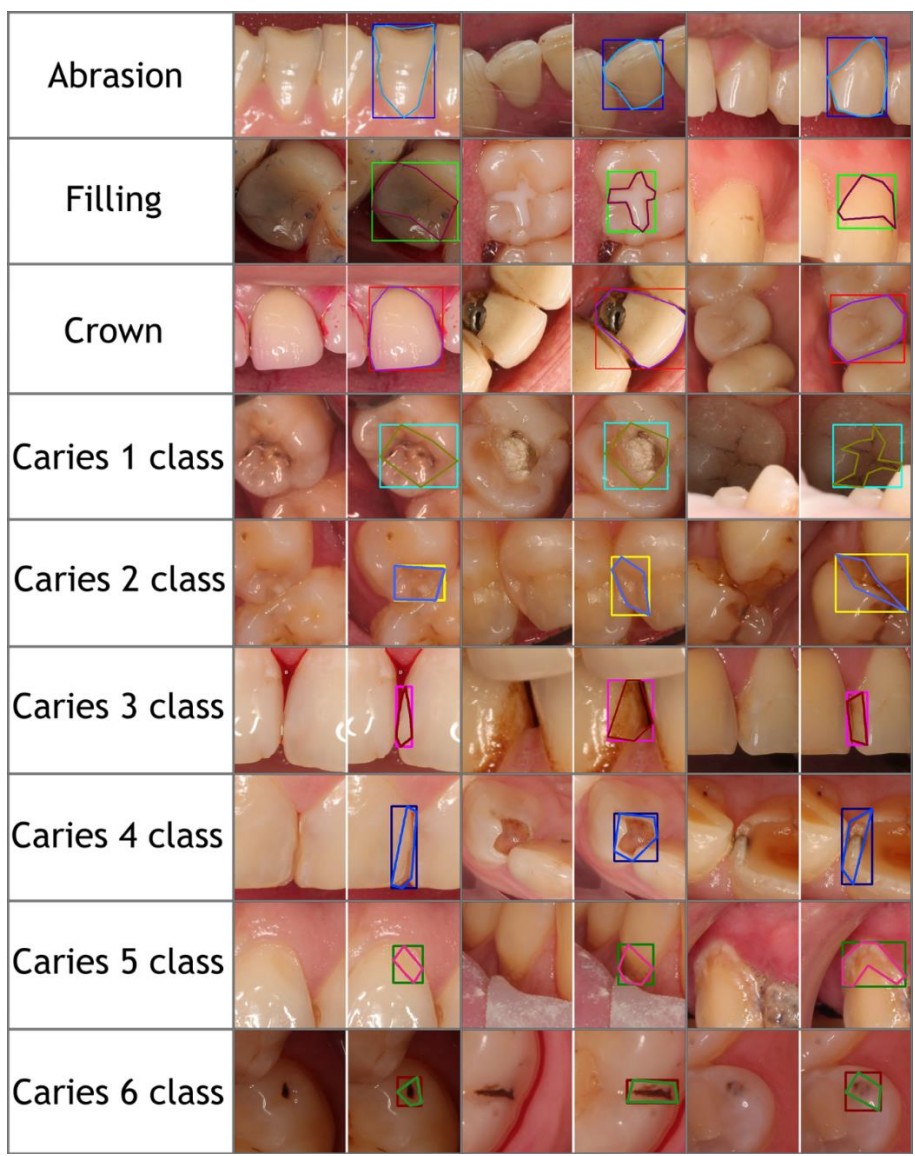

Figure 2. Examples of photo markings for each class

The test data was prepared later than the training data (in May 2025) and labeled separately. In this way, we simulated the process of using the trained models on new data. The test data is not divided by patients but consists of individual images. There are 135 images in the test data. The images themselves are available within the proposed dataset, but the labeling is closed. Closed markup is used on Kaggle Leaderboard [21] for an objective assessment of solutions made by other research teams. The submitted solutions are compared with the standard; the mAP@50 metric is calculated by which the submitted solutions are then ranked.

## 4. DESCRIPTION OF THE DATASET

The prepared dataset was divided into 2 parts: training and validation. The dataset was divided by patients, not by photographs. 273 patients were included in the training part, and 22 patients were included in the validation part. Detailed statistics on classes in training and validation are given in Table 2.

Table 2. Statistics on the distribution of pathology classes according to training and validation data

| Class name | Training data | | Validation data | |
|---|---|---|---|---|
| | Number of images containing the class | Number of marked masks for the class | Number of images containing the class | Number of marked masks for the class |
| Abrasion | 1065 | 5957 | 73 | 409 |
| Filling | 695 | 2187 | 48 | 186 |
| Crown | 210 | 570 | 9 | 19 |
| Caries 1 class | 341 | 742 | 30 | 62 |
| Caries 2 class | 471 | 1026 | 41 | 73 |
| Caries 3 class | 238 | 474 | 23 | 33 |
| Caries 4 class | 30 | 43 | 3 | 4 |
| Caries 5 class | 328 | 981 | 21 | 81 |
| Caries 6 class | 33 | 52 | 4 | 5 |

Dataset statistics:

1) Number of male patients: 131, number of female patients: 162;

2) Average number of images per patient: 4.5;

3) Most images have a resolution of more than 5000 by 3000 (15 megapixels);

4) Average number of rectangles per image: 9.77.

The age distribution of patients is shown in Figure 3.

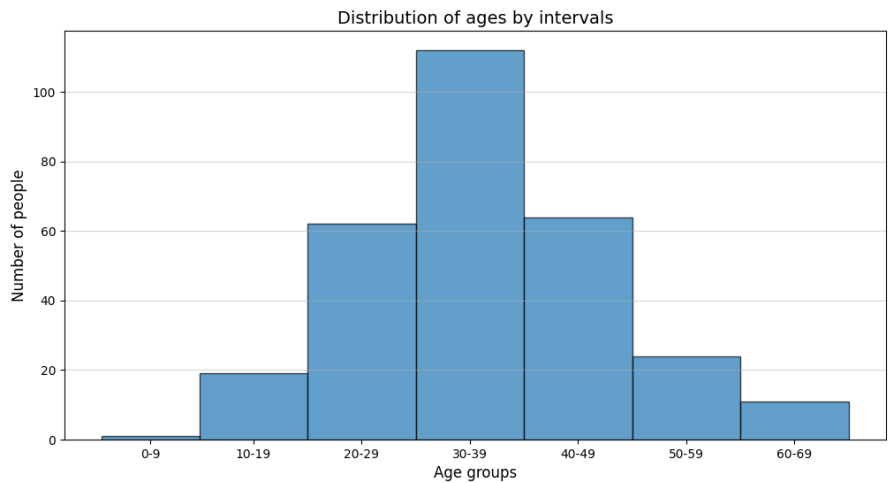

Figure 3. Distribution of patients by age

# 5. DIRECTORY STRUCTURE AND DATA FORMAT

Below is the directory structure of the dataset. It is designed in such a way that it is possible to start training Yolo models [22] without modifications.

```
AlphaDent
--- images - the folder contains all the images of the dataset divided into 3 folders
------ train
--------- p001_F_32_001.jpg
--------- p001_F_32_002.jpg
--------- … (total xxxx images)
------ valid
--------- p001_M_63_001.jpg
--------- p001_M_63_002.jpg
--------- … (total xxxx images)
------ test
--------- test_000.jpg
--------- test_001.jpg
--------- … (total 121 images)
--- labels - the folder contains markup for images in the form of polygons and masks
------ train
--------- p001_F_32_001_masks
------------ 00_class_2.png
------------ 01_class_5.png
------------ … (all N masks for the image 'p001_F_32_001.jpg' are in png format)
--------- p001_F_32_002_masks
------------ 00_class_3.png
------------ 01_class_7.png
------------ … (all K masks for the image 'p001_F_32_002.jpg' are in png format)
--------- … (folders with masks for all training images)
--------- p001_F_32_001.txt
--------- p001_F_32_002.txt
--------- … (text markup for all other training images)
------ valid
--------- p001_M_63_001_masks
------------ 00_class_4.png
------------ 01_class_5.png
------------ … (all M masks for the image 'p001_M_63_001.jpg' are in png format)
--------- p001_M_63_002_masks
------------ 00_class_3.png
------------ 01_class_7.png
------------ … (all K masks for the image 'p001_M_63_002.jpg' are in png format)
--------- … (folders with masks for all validation images)
--------- p001_M_63_001.txt
--------- p001_M_63_002.txt
--------- … (text markup for all other validation images)
--- yolo_seg_train.yaml - YAML-file with class description
```

If the image has the name p001_F_32_001.jpg, then in this name:

1) p001 – patient ID;
2) _F_ – sex of the patient (can be M (male) and F (female));
3) _32_ – patient age;
4) _001 – image ID for this patient.

For masks, the name structure is as follows. For the name: 01_class_5.png:

1) 01_ – mask ID for the given image;
2) _5 – image class (from 0 to 8 inclusive).

If the image is located at images/train/p001_F_32_001.jpg, then the corresponding markup is located at labels/train/p001_F_32_001.txt.

The format of the dataset labels is as follows:

1) One text file per image. Each image in the dataset has a corresponding text file with the same name as the image file and a ".txt" extension.
2) One row per feature. Each row in the text file corresponds to one feature in the image.
3) Feature information per row. Each row contains the following information about the feature instance:
    − Object class index: An integer representing the class of the object (0 to 8; as in Table 1).
    − Object boundary coordinates. Boundary coordinates around the mask area normalized to values from 0 to 1.

The format of one line in the segmentation dataset file is as follows:
<class-index> <x1> <y1> <x2> <y2> ... <xn> <yn>

## 6. EXPERIMENTAL RESULTS

For training, the Yolo v8 neural network, one of the most popular networks for solving Object Detection and Instance Segmentation problems, was used [22] (Figure 4).

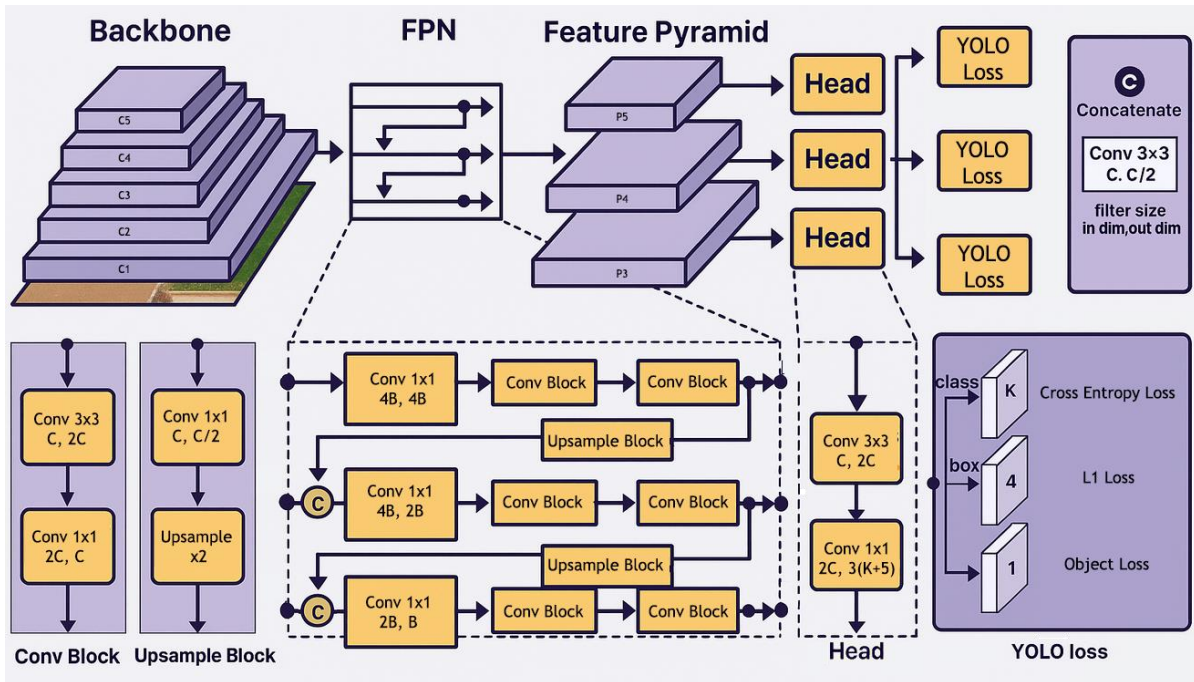

Figure 4.    Yolo v8 neural network architecture

Training settings: 9 classes. Task type: Instance Segmentation. Image sizes are 640px and 960px. Number of epochs: 100. Yolo version: v8x Large set of augmentations. Target metric: average mAP50 across all 9 classes.

The results of validation for the 960px image size are shown in Table 3. For the 640px image size, the mAP@50 value was slightly lower.

Table 3. Experiment results for 9 classes

| Class | Images | Instances | Precision | Recall | mAP@50 | mAP@50:90 |
|---|---|---|---|---|---|---|
| all (average) | 83 | 872 | 0.513 | 0.430 | **0.436** | 0.240 |
| Abrasion | 73 | 409 | 0.630 | 0.878 | 0.768 | 0.638 |
| Filling | 48 | 186 | 0.640 | 0.649 | 0.677 | 0.342 |
| Crown | 9 | 19 | 0.758 | 0.895 | 0.841 | 0.657 |
| Caries 1 class | 30 | 62 | 0.483 | 0.484 | 0.444 | 0.152 |
| Caries 2 class | 41 | 73 | 0.329 | 0.370 | 0.253 | 0.068 |
| Caries 3 class | 23 | 33 | 0.182 | 0.090 | 0.052 | 0.017 |
| Caries 4 class | 3 | 4 | 1.000 | 0.000 | 0.293 | 0.082 |
| Caries 5 class | 21 | 81 | 0.593 | 0.504 | 0.549 | 0.181 |
| Caries 6 class | 4 | 5 | 0.000 | 0.000 | 0.043 | 0.021 |

Designations in the table:

– Images: number of images containing this class;

– Instances: total number of class instances contained in the validation;

– P (Precision): object detection accuracy showing how many detections were correct;

– R (Recall): ability of the model to identify all instances of the objects in the images;

– mAP@50: average accuracy calculated with an intersection over union (IoU) threshold of 0.50. IoU is calculated between the actual mask and predicted mask [23]. This is a measure of the accuracy of the model, taking into account only "easy" detections;

– mAP@50:95: average of the average accuracy calculated at different IoU thresholds ranging from 0.50 to 0.95. This gives a comprehensive view of the model's performance at different levels of the detection difficulty.

All the metrics are between 0 and 1. The closer to 1.0, the more accurate the model is. The target metric that determines the quality of the model is marked in red.

Table 3 shows that there are some types of caries for which there is little data and which are poorly determined by the model, especially for the caries types 3, 4, and 6. In this regard, an additional experiment was conducted in which all 6 types of caries were combined into one class. The results of the experiment are shown in Table 4.

Table 4. Experiment results for 4 classes

| Class | Images | Instances | Precision | Recall | mAP@50 | mAP@50:90 |
|---|---|---|---|---|---|---|
| all (average) | 83 | 872 | 0.626 | 0.690 | 0.680 | 0.458 |
| Abrasion | 73 | 409 | 0.595 | 0.910 | 0.808 | 0.653 |
| Filling | 48 | 186 | 0.642 | 0.696 | 0.701 | 0.366 |
| Crown | 9 | 19 | 0.802 | 0.842 | 0.893 | 0.708 |
| Caries | 64 | 258 | 0.463 | 0.314 | 0.316 | 0.106 |

Examples of the predictions are shown in Figure 5.

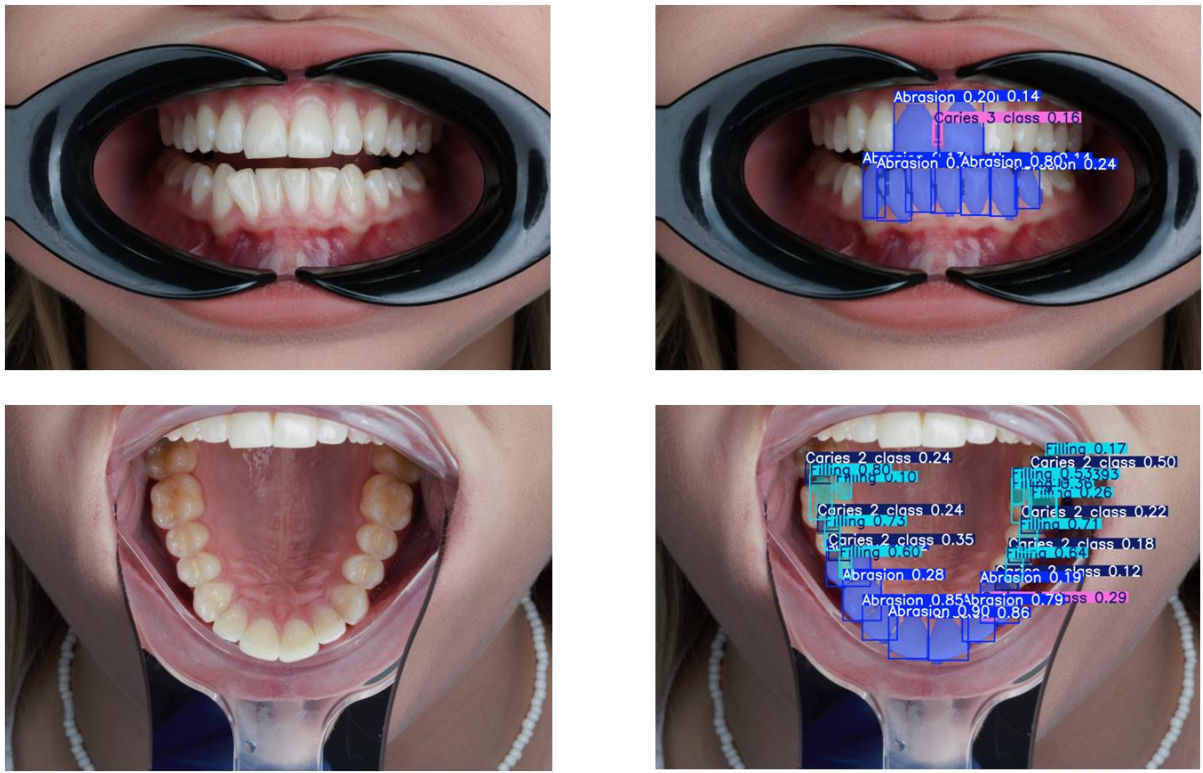

Figure 5. Examples of neural network predictions on validation images

The code we used for training, as well as the code for inference and pre-trained weights, are available here [24]: https://github.com/ZFTurbo/AlphaDent.

## 7. CONCLUSION

Thus, we present a new dental image dataset (AlphaDent). This is a unique dataset based on the DSLR dental photographs and contains over 1200 images for 295 patients. The dataset is fully labeled and divided into 9 classes. The peculiarity of the dataset is that it is published under an open license and is accompanied by the training/inference code and model weights. This makes it indispensable for the research in the field of artificial intelligence for predicting dental diseases. The developed dataset was tested on the Instance Segmentation task and showed high results (up to 0.680 mAP@50).

A potential problem with the dataset is that it was mainly prepared by one specialist. The markup by several specialists will increase the accuracy of the dataset. Also, most of the dataset is collected on one camera, which can affect the accuracy of predictions made from the photographs taken by other cameras. In the future, it is planned to expand the dataset and eliminate the shortcomings described.

## 8. ETHICAL APPROVAL

This study was approved by the Local Ethics Committee of the First Moscow State Medical University named after I.M. Sechenov (protocol No. 02-24 dated 01/29/2024). Written informed consent was obtained from all the participants prior to their inclusion in the study in strict accordance with the standards for conducting, reporting, editing, and publishing biomedical research.

## 9. PUBLICATION

This manuscript has been submitted for publication in Computer Optics journal.